\documentclass[11pt]{article}

\usepackage{acl}

\usepackage{times}
\usepackage{latexsym}

\usepackage[T1]{fontenc}

\usepackage[utf8]{inputenc}

\usepackage{microtype}

\usepackage{inconsolata}

\usepackage{graphicx}
\usepackage{booktabs}
\usepackage{tabularx}
\usepackage{multirow}
\usepackage{amsmath}
\usepackage{subcaption}
\usepackage{amsmath}
\usepackage{amssymb}
\usepackage{makecell}
\usepackage{bm}
\usepackage[table]{xcolor}
\title{Benchmark Leakage Trap: Can We Trust LLM-based Recommendation?}

\author{
  Mingqiao Zhang \\
  Nanjing University \\
  \texttt{231220059@smail.nju.edu.cn}
  \And
  Qiyao Peng \\
  Tianjin University \\
  \texttt{qypeng@tju.edu.cn}
  \And
  Yinghui Wang \\
  Beijing Institute of Control \\ and Electronic Technology \\
  \texttt{wangyinghui@tju.edu.cn} 
  \AND
  Hongtao Liu \\
  Tianjin University \\
  \texttt{htliu@tju.edu.cn}
  \And
  Yumeng Wang \\
  Tianjin University \\
  \texttt{ymwang@tju.edu.cn}
}

\begin{document}
\maketitle
\begin{abstract}
The expanding integration of Large Language Models (LLMs) into recommender systems poses critical challenges to evaluation reliability. This paper identifies and investigates a previously overlooked issue: benchmark data leakage in LLM-based recommendation. This phenomenon occurs when LLMs are exposed to and potentially memorize benchmark datasets during pre-training or fine-tuning, leading to artificially inflated performance metrics that fail to reflect true model performance. To validate this phenomenon, we simulate diverse data leakage scenarios by conducting continued pre-training of foundation models on strategically blended corpora, which include user-item interactions from both in-domain and out-of-domain sources. Our experiments reveal a dual-effect of data leakage: when the leaked data is domain-relevant, it induces substantial but spurious performance gains, misleadingly exaggerating the model's capability. In contrast, domain-irrelevant leakage typically degrades recommendation accuracy, highlighting the complex and contingent nature of this contamination. Our findings reveal that data leakage acts as a critical, previously unaccounted-for factor in LLM-based recommendation, which could impact the true model performance.

\end{abstract}
\section{Introduction}
Recently, the Large Language Models (LLMs) (e.g., GPT-4~\cite{achiam2023gpt}, LLaMA~\cite{touvron2023llama}) demonstrate unprecedented capabilities in natural language understanding, which has been employed in various fields. 
The recommender systems based on large language models (LLMs) have emerged as a promising direction~\cite{wu2024survey,wang2024towards}. 
These models are pre-trained on massive corpora and encode rich knowledge information, which could enhance the recommender systems by deeper understanding of item content, user intent and contextual nuances~\cite{zhang2023recommendation,li2024large,liu2023pre}.

For example, BinLLM~\cite{zhang2024textlikeencodingcollaborativeinformation} integrates collaborative information into LLMs using a text-like encoding
strategy for recommendation, which has demonstrated superior performance.

However, \textbf{should the results of these LLM-based recommender systems really be considered reliable?}
Recent studies have revealed that large language models are susceptible to data leakage, where models inadvertently memorize and reproduce training data during inference~\cite{carlini2021extracting,hisamoto2020membership,carlini2022quantifying}. 
This phenomenon has been extensively documented in various domains, from text generation~\cite{carlini2022extracting} to question answering~\cite{kandpal2022deduplicating}, where LLMs have been shown to reproduce verbatim passages from their training corpora. 
And it largely affects the evaluation of model performance~\cite{zhou2023dontmakellmevaluation,golchin2024time}. 
Moreover, LLM has been proven to have the ability to remember recommendation data~\cite{di2025llms}.
Although there is ample evidence that LLMs are exposed to benchmark data during pre-training and affect downstream inference results\cite{carlini2021extracting,kandpal2022deduplicating}, it remains unclear how leaked LLMs affect downstream recommendation results when used as the backbone model of recommendation systems.

This raises our guess: integrating a memorization-prone LLM into the recommendation pipeline may cause the resulting LLM-based recommender systems to inherit and potentially amplify the data leakage characteristics inherent in the LLM.  
In fact, the core of recommender systems is user interest modeling and item characteristics learning~\cite{zhang2024textlikeencodingcollaborativeinformation,rendle2012factorization}.
However, the inherent data leakage in LLMs can obscure the boundary between authentic user preferences and memorized data artifacts. 
This interference not only distorts the modeling of latent user interests, but also compromises the learning of item representations, as the model may rely on memorized associations rather than underlying and real characteristics. 
Such leakage could ultimately bias the evaluation of LLM-based recommender systems, as test results may reflect the model's prior exposure to evaluation data rather than its true recommendation ability.

To investigate the impact of data leakage on LLM-based recommendation systems, we design an experimental framework that could simulate real-world leakage scenarios. 
First, we construct mixed leakage datasets with controlled proportions by blending in-domain data (sampled from the target evaluation datasets) with out-of-domain data (collected from six external sources).
Second, to simulate leakage scenarios,  we fine-tune the LLMs via Low-Rank Adaptation (LoRA) based on the blended leakage dataset, named as dirty LLM. This choice of LoRA provides a controlled proxy for investigating how benchmark knowledge injection biases recommendation performance while preserving the foundation model's base capabilities. 
Finally, we evaluated performance by comparing downstream recommendation models built on clean and dirty LLMs in two categories. 
Specifically, we evaluate the impact of data leakage on (1) different model architectures, (2) various leakage compositions (pure in-domain, pure out-of-domain, and mixed), and (3) different types of out-of-domain data sources. 
In this way, we can analyze the factors that determine a model's susceptibility or robustness to data leakage.

Our work makes the following key contributions:
\begin{enumerate}
    \item We identify and empirically demonstrate the benchmark data leakage problem in LLM-based recommender systems, revealing how pre-exposed LLMs can compromise the integrity of downstream evaluation metrics.
    \item We develop a novel methodology for simulating realistic data leakage scenarios through controlled fine-tuning on strategically composed mixed-domain datasets, enabling investigation of how different types and degrees of data exposure affect recommendation performance.
    \item Through extensive experiments, we demonstrate that: (i) impacts vary substantially across different model architectures and their resilience to leakage, (ii) in-domain leakage creates deceptive performance gains that mask true model capability, while (iii) out-of-domain contamination typically impairs performance, revealing a nuanced dual-effect phenomenon that challenges current evaluation practices.
\end{enumerate}

\section{Related Work}

\subsection{Large Language Models}
Large Language Models (LLMs) are a class of foundational models pre-trained on vast corpora of text data, demonstrating remarkable capabilities in understanding and generating human-like text~\cite{achiam2023gpt,bommasani2021foundation,brown2020language}. These models, typically based on the Transformer architecture, learn to predict the next token in a sequence, thereby capturing intricate patterns, facts, and reasoning abilities embedded in the training data~\cite{wei2022emergent,kaplan2020scaling}. 
A significant security and privacy concern associated with these models is 
\textbf{data leakage}, wherein LLMs can store and subsequently reproduce verbatim content from their training datasets~\cite{carlini2021extracting}. 
The study of data leakage in LLMs has become a critical area of research. Seminal work by Carlini et al.~
\cite{carlini2021extracting} first systematically demonstrates that LLMs can memorize and regurgitate training data, showing that simple prompting strategies could recover exact sequences. ~\cite{zhou2023dontmakellmevaluation} shows that data leakage of large models will interfere with subsequent performance evaluation. And in response to address this problem, mitigation strategies like differential privacy and training data deduplication have been proposed to reduce these risks~\cite{kandpal2022deduplicating}.

\subsection{LLM-based Recommendation}

The field of LLM-based recommendation has evolved rapidly, with researchers exploring various architectural paradigms to integrate language models with recommendation tasks~\cite{wu2024survey,lin2025can}. LLMRec and LLMRec+Collab. are two primary architectural paradigms, which are directly represented in our experimental evaluation. 

The first category, \textbf{LLMRec methods}, leverages the inherent language understanding capabilities of LLMs directly, with minimal architectural modification. This category encompasses a spectrum of techniques: ICL~\cite{dai2023can} operates in a zero-shot manner through in-context learning alone, Prompt4NR~\cite{zhang2023prompt} explores generic soft-prompt tuning for the recommendation task, while TALLRec~\cite{bao2023tallrec} specializes the model via instruction tuning. 

In contrast, the second category, \textbf{LLMRec+Collab. methods}, explicitly integrates collaborative filtering signals to enhance the LLM's reasoning with user-item interaction patterns. The integration strategies, however, vary significantly: Personalized recommendation~\cite{lyu2024llm} injects collaborative information by introducing trainable token embeddings for users and items, effectively performing personalized soft-prompt tuning. CoLLM~\cite{10882951}, in its two instantiations, maps pre-trained collaborative embeddings from Matrix Factorization (CoLLM-MF) or a Deep Interest Network (CoLLM-DIN) into the LLM's latent space. A distinctly different approach is taken by BinLLM~\cite{zhang2024textlikeencodingcollaborativeinformation}, which encodes collaborative signals into a text-like format of binary hash codes for direct inclusion in the model's prompt. This clear taxonomic distinction provides the essential context for interpreting our experimental results on how these fundamentally different paradigms respond to data leakage scenarios.

\section{Methodology}
This section details our experimental framework for investigating  data leakage in LLM-based recommender systems. We first provide an overview of the experimental pipeline, followed by the construction of leakage data, the contamination process through LoRA adaptation, and the formulation of clean and dirty downstream recommenders.

\begin{figure*}[t] 
\centering  
\includegraphics[width=\textwidth]{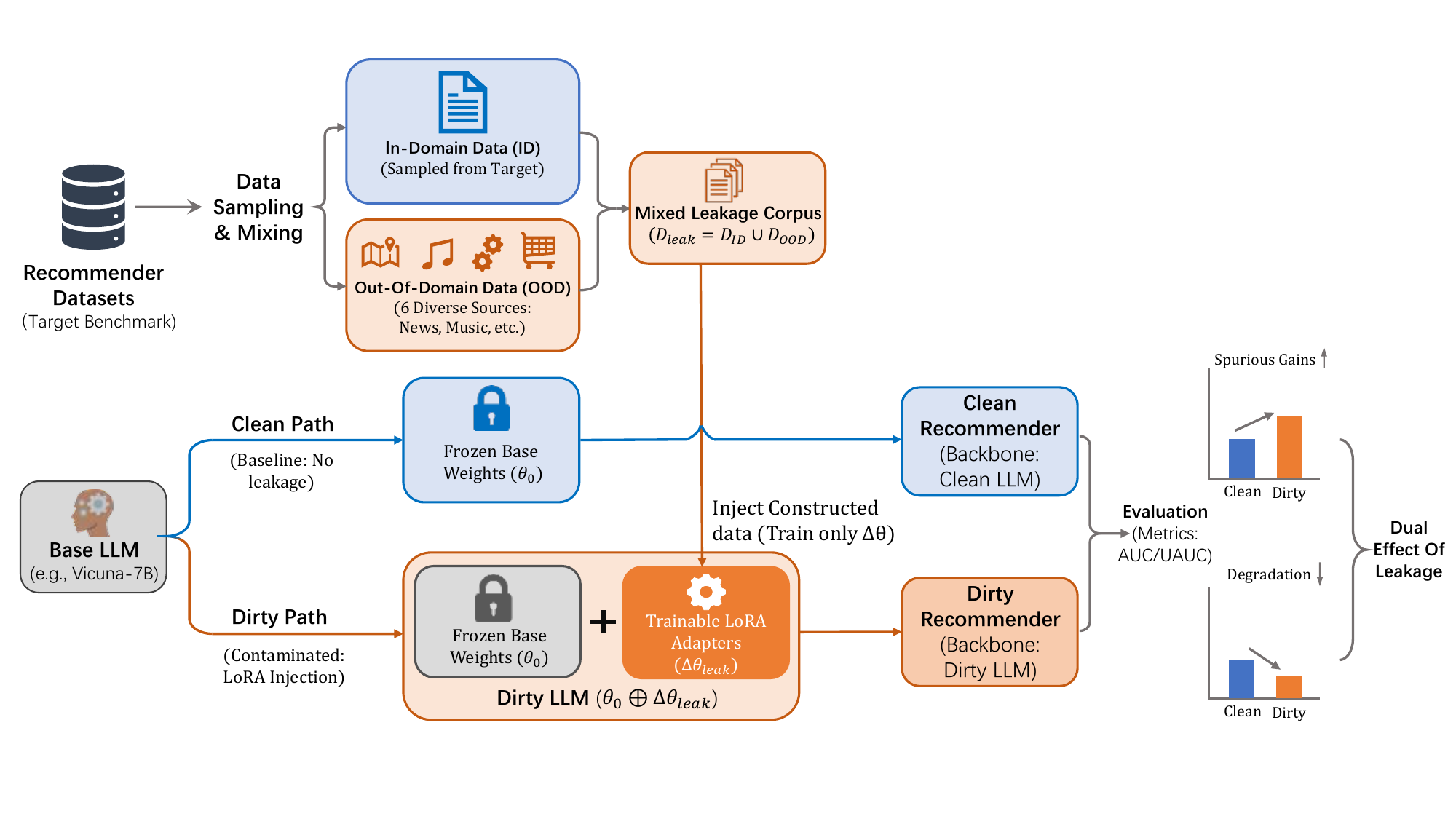}  
\vspace{-12mm}
\caption{Overview of the Experimental Framework. The workflow illustrates the construction of a Mixed Leakage Corpus through strategic sampling of In-Domain (ID) and Out-Of-Domain (OOD) data. The Clean Path represents the baseline evaluation using a frozen Base LLM, while the Dirty Path simulates benchmark contamination via LoRA injection. By comparing the Clean and Dirty Recommenders, we identify a dual-effect of leakage: spurious performance gains or degradation.}    
\label{fig:Process of the experiments.}
\vspace{-4mm}
\end{figure*}

\subsection{Overview}
The complete workflow of our study is illustrated in Figure~\ref{fig:Process of the experiments.}. To explore the effect of data leakage on recommendation models, we begin by sampling a portion of the original dataset to serve as leaked data, which contains a pre-defined ratio of in-domain (ID) and out-of-domain (OOD) samples depending on whether the data originate from the same domain as the evaluation benchmark.

We start from an open-source large language model (Vicuna-7B) that has already been pre-trained on large-scale text corpora and denote it as the Clean LLM (i.e., without leakage). To emulate contamination, we fine-tune this Clean LLM on the mixed leakage dataset using Low-Rank Adaptation (LoRA)~\cite{hu2022lora}. Through this process, the model learns leakage-related patterns that are encoded into the newly added low-rank parameters, producing a Dirty LLM. The base parameters remain frozen throughout training, ensuring that only the LoRA modules carry the injected information.

Subsequently, both Clean and Dirty LLMs are employed as backbone encoders in downstream recommendation systems, forming Clean Recommenders and Dirty Recommenders respectively. All hyperparameters, data partitions, and model configurations are kept identical between the two to ensure a fair comparison. The performance differences between the two systems therefore directly reflect the influence of data leakage.

During evaluation, Clean Recommenders serve as baseline models for standard performance testing, while Dirty Recommenders are used to observe the potential distortion of benchmark results under contaminated conditions. By comparing the performance metrics (AUC/UAUC) of the Clean and Dirty Recommenders, we identify the \textit{Dual Effect of Leakage}: (1) \textbf{Spurious Gains} induced by domain-relevant leakage and (2) performance \textbf{Degradation} caused by domain-irrelevant contamination. This design allows us to analyze how different leakage types and proportions affect model reliability in a controlled and reproducible way.

\subsection{Leakage Data Construction}
To simulate realistic benchmark leakage scenarios, we construct a mixed leakage dataset that combines both in-domain and out-of-domain sources. Each dataset is split into training, validation, and test sets following a 7:2:1 ratio.

First, we randomly sample 10\% from the target dataset as in-domain leakage:
\begin{equation}
D_{\mathrm{ID}} \sim \mathrm{Sample}\!\left(D_{\text{target}},\, p=0.1\right).
\label{eq:id}
\end{equation}
Here, $D_{\mathrm{ID}}$ denotes the in-domain leakage subset drawn from the target dataset $D_{\text{target}}$ at proportion $p{=}0.1$. The operator $\mathrm{Sample}(\cdot)$ denotes uniform sampling according to the stated proportion.
\textbf{$D_{\mathrm{ID}}$ contains training set, validation set, and test set, and the ratio remains 7:2:1.}
This step fixes the reference scale for subsequent OOD construction.

Then, we draw OOD data from six external datasets (Epinions, Last.fm, MIND, Amazon\text{-}Sports, Amazon\text{-}Beauty, Gowalla), with the total OOD size equal to $6\times |D_{\mathrm{ID}}|$ and allocated equally across sources:
\begin{equation}
D_{\mathrm{OOD}}^{(i)} \sim \mathrm{Sample}\!\left(D_{\text{ext}}^{(i)},\, |D_{\mathrm{ID}}|\right),\quad i\in\{1,\dots,6\},
\label{eq:ood_i}
\end{equation}
\begin{equation}
D_{\mathrm{OOD}} \;=\; \bigcup_{i=1}^{6} D_{\mathrm{OOD}}^{(i)}, 
\qquad |D_{\mathrm{OOD}}| \;=\; 6\,|D_{\mathrm{ID}}|.
\label{eq:ood_union}
\end{equation}
In \eqref{eq:ood_i}, each $D_{\mathrm{OOD}}^{(i)}$ is sampled from the $i$-th external dataset $D_{\text{ext}}^{(i)}$ with \emph{the same} target size $|D_{\mathrm{ID}}|$, ensuring that every source contributes equally. The union in \eqref{eq:ood_union} aggregates the six OOD portions into $D_{\mathrm{OOD}}$, yielding a total size of $6\,|D_{\mathrm{ID}}|$. Here $|\cdot|$ denotes cardinality.

Finally, we combine ID and OOD to obtain the mixed leakage set:
\begin{equation}
\begin{split}
D_{\text{leak}} &= D_{\mathrm{ID}} \cup D_{\mathrm{OOD}}, \\
|D_{\text{leak}}| &= |D_{\mathrm{ID}}| + |D_{\mathrm{OOD}}| = 7\,|D_{\mathrm{ID}}|.
\end{split}
\label{eq:leak}
\end{equation}
Equation \eqref{eq:leak} defines the final leakage corpus $D_{\text{leak}}$ as the combination of the in-domain portion $D_{\mathrm{ID}}$ and the aggregated out-of-domain portion $D_{\mathrm{OOD}}$, whose size follows directly from the construction above, i.e., $|D_{\text{leak}}|=7\,|D_{\mathrm{ID}}|$.

$D_{\text{leak}}$ is transformed into prompts and then injected into the LLM. 
\textbf{Appendix~\ref{app:prompts} shows the details.}

\subsection{Injecting leaked data}
We start from an already pretrained base LLM with parameters \(\theta_{0}\).
This model is used directly as the Clean LLM and its parameters remain
frozen throughout the whole study. To obtain the Dirty LLM, we
optimize only low-rank LoRA adapters on the mixed leakage dataset
\(D_{\text{leak}}\) (constructed in~\eqref{eq:leak}), while keeping
\(\theta_{0}\) unchanged. 

\paragraph{Loss definition}
Let \(L_{\text{LLM}}(\cdot)\) denote the standard negative log-likelihood (NLL)
under next-token prediction:
\begin{equation}
L_{\text{LLM}}(\theta, D)
= - \sum_{(x,y)\in D} \log P(y \mid x; \theta),
\label{eq:llm-loss}
\end{equation}
where each pair \((x,y)\) consists of an input sequence \(x\) and target tokens \(y\),
and \(P(y\mid x;\theta)\) is the model likelihood parameterized by \(\theta\).
This loss is adopted \emph{only} to train LoRA adapters; it does not modify the
frozen base \(\theta_0\).

\paragraph{Contaminated adaptation (Dirty LLM)}
We train LoRA adapters on \(D_{\text{leak}}\) as follows:
\begin{equation}
\Delta\theta^{\text{leak}}_{\text{LoRA}}
= \arg\min_{\Delta\theta_{\text{LoRA}}}
\;L_{\text{LLM}}\!\big(\theta_{0} \oplus \Delta\theta_{\text{LoRA}},\; D_{\text{leak}}\big),
\label{eq:lora-opt}
\end{equation}
\begin{equation}
\theta^{\text{dirty}}_{\text{LLM}}
= \theta_{0} \oplus \Delta\theta^{\text{leak}}_{\text{LoRA}}.
\label{eq:dirty-llm}
\end{equation}
Here, \(\oplus\) denotes the \emph{injection} of LoRA adapters into the frozen base
(i.e., composing the base weights with low-rank additive updates).\footnote{We
use the same tokenizer and decoding head as the base model; only LoRA
parameters are trainable.} The optimization in~\eqref{eq:lora-opt} finds
the adapter parameters that \emph{best fit the leakage corpus}, thereby encoding
leakage-specific patterns into \(\Delta\theta^{\text{leak}}_{\text{LoRA}}\).
Equation~\eqref{eq:dirty-llm} defines the Dirty LLM as the composition of the
unchanged base \(\theta_0\) and the learned leakage adapters.

\paragraph{LoRA parameterization and what is updated}
For any affine weight \(W\) in the frozen base, the LoRA update is
\begin{equation}
W_{\text{new}} = W + \frac{\alpha}{r}\, BA,
\label{eq:lora-param}
\end{equation}
where \(B\in\mathbb{R}^{d\times r}\) and \(A\in\mathbb{R}^{r\times k}\) are trainable
low-rank matrices (rank \(r\ll \min(d,k)\)), and \(\alpha\) scales the update
magnitude. The product \(BA\) is a rank-\(r\) correction that \emph{adds} to
the frozen weight \(W\); hence
\emph{only} \(A,B\) are optimized, while all base weights \(W\subset\theta_0\)
remain fixed. This factorization provides two benefits: (i) \textbf{parameter
efficiency} (few trainable parameters) and (ii) \textbf{isolation of contamination}
(leakage information is confined within \(\Delta\theta_{\text{LoRA}}\), enabling a clean
comparison against the base model).

In subsequent experiments, the Clean LLM is simply \(\theta_0\) (no adapters),
and the Dirty LLM is \(\theta^{\text{dirty}}_{\text{LLM}}\) from \eqref{eq:dirty-llm}. 
\textbf{And the reason for using LoRa fine-tuning instead of full-parameter training is explained in Appendix~\ref{app:Justification for LoRA}.}

\begin{table*}[ht]
\small
\setlength{\tabcolsep}{4pt} 
\renewcommand{\arraystretch}{1.5}
\centering
\resizebox{\linewidth}{!}{
\begin{tabular}{c|c|c| c c c  | c c c}
\toprule
\multirow{2}{*}{\textbf{Dataset}} & 
\multirow{2}{*}{\textbf{Method}} & 
\multirow{2}{*}{\textbf{Category}} & 
\multicolumn{3}{c|}{\textbf{AUC}} & 
\multicolumn{3}{c}{\textbf{UAUC}} \\
\cmidrule(lr){4-6} \cmidrule(lr){7-9}
& & & 
\textbf{Baseline} & \textbf{After leak} & $\mathbf{\Delta}$\textbf{AUC} & 
\textbf{Baseline} & \textbf{After leak} & $\mathbf{\Delta}$\textbf{UAUC} \\
\midrule
\multirow[c]{8}{*}{\parbox[c]{1.4cm}{\centering \textbf{ML-} \\[4pt] \textbf{1M}}}
& ICL           & \multirow{3}{*}{LLMRec}          & 0.5621 & 0.5034 & $\bm{-10.4\%}$ & 0.5518 & 0.5015 & $\bm{-9.1\%}$ \\
& Prompt4NR     &                                  & 0.7021 & 0.6515 & $-7.2\%$  & 0.6679 & 0.6180 & $-7.5\%$  \\
& TALLRec       &                                  & 0.7036 & 0.6215 & $\bm{-11.7\%}$ & 0.6751 & 0.5902 & $\bm{-12.6\%}$ \\
\cmidrule(lr){2-9}
& CoLLM-MF      & \multirow{4}{*}{\makecell{LLMRec \\ +Collab.}} & 0.7228 & 0.7685 & $+6.3\%$ & 0.6782 & 0.7251 & $+6.9\%$ \\
& CoLLM-DIN     &                                  & 0.7184 & 0.7512 & $+4.6\%$  & 0.6801 & 0.7085 & $+4.2\%$  \\
& PersonPrompt  &                                  & 0.7151 & 0.6914 & $-3.3\%$  & 0.6487 & 0.6275 & $-3.3\%$  \\
& BinLLM        &                                  & 0.7379 & 0.7823 & $+6.0\%$  & 0.6908 & 0.7345 & $+6.3\%$  \\
\midrule
\multirow[c]{8}{*}{\parbox[c]{1.6cm}{\centering \textbf{Amazon-} \\[4pt] \textbf{Book}}}
& ICL           & \multirow{3}{*}{LLMRec}          & 0.5542 & 0.5011 & $\bm{-9.6\%}$ & 0.5487 & 0.5008 & $\bm{-8.7\%}$ \\
& Prompt4NR     &                                  & 0.7168 & 0.6275 & $\bm{-12.5\%}$ & 0.5804 & 0.5052 & $\bm{-13.0\%}$ \\
& TALLRec       &                                  & 0.7302 & 0.7758 & $+6.2\%$  & 0.5892 & 0.6285 & $+6.7\%$  \\
\cmidrule(lr){2-9}
& CoLLM-MF      & \multirow{4}{*}{\makecell{LLMRec \\ +Collab.}} & 0.8034 & 0.8351 & $+3.9\%$ & 0.6147 & 0.6415 & $+4.4\%$ \\
& CoLLM-DIN     &                                  & 0.8186 & 0.7814 & $-4.5\%$  & 0.6401 & 0.6135 & $-4.2\%$  \\
& PersonPrompt  &                                  & 0.7209 & 0.7712 & $+7.0\%$  & 0.5876 & 0.6251 & $+6.4\%$  \\
& BinLLM        &                                  & 0.8207 & 0.7456 & $\bm{-9.1\%}$  & 0.6262 & 0.5723 & $\bm{-8.6\%}$  \\
\bottomrule
\end{tabular}%
}
\vspace{3pt}
\caption{AUC/UAUC comparison on the ML-1M and Amazon-Book datasets under mixed data leakage (in-domain 10\% + out-of-domain 60\%). “Collab.” denotes collaborative recommendation methods. $\Delta$AUC denotes the rate of change of AUC. $\Delta$UAUC denotes the rate of change of UAUC. Bold values in $\Delta$ columns indicate significant performance variations ($|\Delta| \geq 8\%$).}
\label{tab:llm_leakage_both}
\vspace{-4mm}
\end{table*}

\section{Experiment}
In this section, we first describe the overall experimental settings, including datasets and baselines. Then, we present and analyze the experimental results to evaluate the effectiveness of our proposed method. 
\textbf{The implementation details are provided in Appendix \ref{details}.}

\subsection{Experiment settings}

\subsubsection{Datasets}
For evaluating the impact of LLM data leakage on downstream recommendation models, we conduct experiments on two widely-used recommendation datasets:
\begin{itemize}
    \item \textbf{ML-1M}~\cite{harper2015movielens}:MovieLens-1M recommendation benchmark, a widely used dataset for movie recommendation containing 1000209 timestamped user–movie ratings (1–5 scale) from 6040 users on 3952 movies. In addition to user-item interactions, it provides side information such as movie titles, etc.
    \item \textbf{Amazon-Book}~\cite{ni2019justifying}: This dataset is derived from the "Books" category of the well-known Amazon Product Review dataset. It comprises book purchasing and rating records, user-book interactions within the e-commerce domain.
\end{itemize}

For simulating the data leakage, we employ six external datasets from different domains to serve as out-of-domain leakage data, including Epinions~\cite{massa2007trust}, Last.fm~\cite{celma2009music}, MIND~\cite{wu2020mind}, Amazon-Sports~\cite{ni2019justifying}, Amazon-Beauty~\cite{ni2019justifying} and Gowalla~\cite{cho2011friendship}. 
\textbf{The detailed information of the above datasets is presented in Appendix ~\ref{statistics}.}

These OOD datasets are strategically selected based on two criteria: 
(i) \textit{Semantic Divergence}: they cover diverse domains ranging from news to geographical check-ins to test the model's resistance to non-target semantic noise; 
(ii) \textit{Structural Heterogeneity}: they encompass varied interaction types (e.g., tags, attributes) to observe how the memorization of disparate behavioral patterns interferes with target recommendation logic.

\subsubsection{Baselines and Metrics}
In this paper, we evaluate the model performance on two categories of LLM-based recommendation models: LLMRec models (ICL, Prompt4NR, TALLRec) and LLMRec models integrated collaborative information (PersonPrompt, CoLLM, BinLLM).
\textbf{Detailed descriptions of all baseline configurations are provided in Appendix \ref{baseline} and the metrics of our experiment are provided in Appendix \ref{app:metrics}.}

\subsection{Results and Analysis}
\subsubsection{Overall Experimental Results} We present the performance comparison of seven recommendation methods on ML-1M and Amazon-Book datasets under mixed data leakage (in-domain 10\% + out-of-domain 60\%) in Table~\ref{tab:llm_leakage_both}. Overall, data leakage can affect the evaluation of model performance to some extent. We have the following observations.

First, the impact of data leakage on model performance is uncertain. Compared to the baseline, the recommender's performance on the two datasets sometimes improves and sometimes deteriorates. In the ML-1M dataset, TALLRec's AUC drops from 0.7036 to 0.6251 after injecting mixed data. And CoLLM-MF's AUC upgrades from 0.7228 to 0.7685. The same model can perform differently on different datasets. For example, BinLLM upgrades from 0.7379 to 0.7823 but it drops from 0.8207 to 0.7456. It is also worth noting that the trends and magnitudes of AUC and UAUC are basically consistent. This indicates that data leaks have a roughly consistent impact on both overall and user-level recommendation tasks. So in the subsequent analysis, we only refer to the changes in AUC.

Second, the impact of data leakage on model performance evaluation cannot be ignored. Recommenders trained on LLMs with data leaks may change their original rankings when undergoing benchmark testing. In the Amazon-book dataset, PersonPrompt's performance is not as good as BinLLM. But after injecting the mixed dataset, PersonPrompt is superior to BinLLM under this benchmark.

Finally, different categories of models have varying sensitivities to data leakage. Recommendation models of the "LLMRec" class (Uses a large language model directly for recommendation) are more susceptible to the effects of hybrid databases compared with recommendation models of "LLMRec+Collab." class (Augments an LLM with collaborative filtering signals). LLMs are strong at modeling textual semantics, whereas recommendation relies heavily on collaborative signals from user–item interactions. Injecting or fine-tuning LLMs with collaborative information aligns them with the behavior-prediction objective, leading to better personalization, ranking accuracy, and robustness.
Therefore, on the baseline, LLMRec+Collab. models perform better. 
After injecting hybrid data, their performance change is smaller than that of pure LLMRec models. 
We speculate that this is because pure LLMRec methods lack alternative signals to offset contaminated knowledge, 
making them more vulnerable to degradation. 
By contrast, LLMRec+Collab. methods integrate collaborative information as an additional signal that provides redundancy, cross-signal validation and robustness enhancement, yielding a system that is less dependent on potentially leaked LLM parameters.
\subsubsection{Proof of Memorization}
A critical question arises from the observed spurious gains under in-domain leakage: does the performance inflation stem from the model genuinely learning domain-specific collaborative patterns, or is it merely remembering pre-exposed targets? 
To definitively disentangle these two mechanisms, we conducted a rigorous overlap auditing analysis. 
We partitioned the test set into two mutually exclusive subsets: the Seen subset, comprising user-item interaction pairs that explicitly appeared in the injected leakage corpus, and the Unseen subset, containing completely novel interactions.

\begin{table}[t]
\centering
\small
\setlength{\tabcolsep}{0pt} 
\begin{tabular*}{\columnwidth}{@{\extracolsep{\fill}}lll ccc}
\toprule
\textbf{Dataset} & \textbf{Metric} & \textbf{Subset} & \textbf{Clean} & \textbf{Dirty} & $\Delta$ \\
\midrule
\multirow{4}{*}{ML-1M} 
& \multirow{2}{*}{AUC}   & Seen   & 0.7412 & 0.9587 & \textbf{+29.3\%} \\
&                        & Unseen & 0.7368 & 0.7556 & \textbf{+2.6\%} \\
\cmidrule{2-6}
& \multirow{2}{*}{UAUC} & Seen   & 0.6897 & 0.8784 & \textbf{+27.4\%} \\
&                        & Unseen & 0.6923 & 0.7187 & \textbf{+3.8\%} \\
\midrule
\multirow{4}{*}{Amazon-Book} 
& \multirow{2}{*}{AUC}   & Seen   & 0.8231 & 0.9143 & \textbf{+11.1\%} \\
&                        & Unseen & 0.8194 & 0.8276 & \textbf{+1.0\%} \\
\cmidrule{2-6}
& \multirow{2}{*}{UAUC} & Seen   & 0.6304 & 0.7437 & \textbf{+18.0\%} \\
&                        & Unseen & 0.6248 & 0.6395 & \textbf{+2.4\%} \\
\bottomrule
\end{tabular*}
\caption{Performance comparison of the BinLLM architecture on memorized (\textit{Seen}) vs. novel (\textit{Unseen}) interaction pairs across ML-1M and Amazon-Book datasets under 10\% in-domain leakage.}
\label{tab:overlap_audit_binllm}
\end{table}

The quantitative results, presented in Table \ref{tab:overlap_audit_binllm}, reveal a stark contrast. Under the 10\% in-domain leakage setting, the Dirty LLM exhibits an extraordinary performance surge exclusively on the Seen subset. 
For instance, on ML-1M, the AUC and UAUC inflate by +29.3\% and +27.4\%, respectively. Conversely, on the Unseen subset, these gains violently diminish to a marginal +2.6\% and +3.8\%. 
A consistent phenomenon is observed in the Amazon-Book dataset. 
This indicates that the performance improvement of the model is extremely largely based on the subset of tests it has seen. 
The model essentially improves performance through memory retrieval rather than through strong generalization ability.
\textbf{Appendix~\ref{app:qualitative case studies} shows further qualitative case studies.}

\subsubsection{The Impact of Leakage Intensity and Domain Divergence} 
We also randomly generate a few other mixed datasets to simulate possible data leaks. In the following experiment, to save space, we just represent the performance of TALLRec, CoLLM-MF and BinLLM under the baseline ML-1M. Note that Tables 3–6 use independently re-sampled leakage corpora and therefore do not need to numerically reconcile with Table~\ref{tab:llm_leakage_both}. We report them to illustrate qualitative trends across leakage compositions rather than to compare absolute magnitudes.

\begin{table}[ht] 
\centering
\renewcommand{\arraystretch}{1.08}
\setlength{\tabcolsep}{2pt} 
\resizebox{\columnwidth}{!}{ 
\begin{tabular}{c | c c | c c}
\toprule
\multicolumn{5}{c}{\textbf{ML-1M (pure ID interference: 10\% ID data)}}\\
\midrule
\multicolumn{1}{c|}{\textbf{Metric}} & \multicolumn{2}{c|}{\textbf{AUC}} & \multicolumn{2}{c}{\textbf{UAUC}} \\
\cmidrule(lr){1-1}\cmidrule(lr){2-3}\cmidrule(lr){4-5}
\textbf{Method} & \textbf{After leak} & \textbf{$\Delta$AUC} &
\textbf{After leak} & \textbf{$\Delta$UAUC} \\
\midrule
TALLRec  & 0.8795 & $+25.0\%$ & 0.8474 & $+25.5\%$ \\
CoLLM-MF & 0.8234 & $+13.9\%$ & 0.7723 & $+13.9\%$ \\
BinLLM   & 0.8456 & $+14.6\%$ & 0.7923 & $+14.7\%$ \\
\bottomrule
\end{tabular}
}
\vspace{6pt}
\caption{AUC/UAUC after leak and relative change rate $\Delta$ on ML-1M under the “LLM stage 10\% data leakage” setting. ID denotes the In-Domain.}
\label{tab:ml1m_llm_10pct_header}
\end{table}

\begin{table}[ht] 
\centering
\renewcommand{\arraystretch}{1.08}
\setlength{\tabcolsep}{2pt} 
\resizebox{\columnwidth}{!}{ 
\begin{tabular}{c | c c | c c}
\toprule
\multicolumn{5}{c}{\textbf{ML-1M (pure OOD interference: 60\% OOD data)}}\\
\midrule
\multicolumn{1}{c|}{\textbf{Metric}} & \multicolumn{2}{c|}{\textbf{AUC}} & \multicolumn{2}{c}{\textbf{UAUC}} \\
\cmidrule(lr){1-1}\cmidrule(lr){2-3}\cmidrule(lr){4-5}
\textbf{Method} & \textbf{After leak} & \textbf{$\Delta$AUC} &
\textbf{After leak} & \textbf{$\Delta$UAUC} \\
\midrule
TALLRec  & 0.5234 & $-25.6\%$ & 0.4923 & $-27.1\%$ \\
CoLLM-MF & 0.6345 & $-12.2\%$ & 0.5923 & $-12.7\%$ \\
BinLLM   & 0.6234 & $-15.5\%$ & 0.5823 & $-15.7\%$ \\
\bottomrule
\end{tabular}
}
\vspace{6pt}
\caption{AUC/UAUC after leak and relative change rate $\Delta$ on ML-1M under the “pure OOD 60\%” setting. OOD denotes the Out-of-Domain.}
\label{tab:ml1m_pure_ood_60_header}
\end{table}

In Tables ~\ref{tab:ml1m_llm_10pct_header} and ~\ref{tab:ml1m_pure_ood_60_header}, we study purely data leakage by only injecting ID (In-Domain) data or OOD (Out-of-Domain), which are two idealized situations. It can be observed that after a data breach, the performance of models in Table ~\ref{tab:ml1m_llm_10pct_header} improves, while the performance of models in Table ~\ref{tab:ml1m_pure_ood_60_header} declines. This indicates that data leakage within the domain can improve the accuracy of model predictions, whereas the opposite is true for out-of-domain data.

\begin{table}[ht] 
\centering
\renewcommand{\arraystretch}{1.08}
\setlength{\tabcolsep}{2pt} 
\resizebox{\columnwidth}{!}{ 
    \begin{tabular}{c | c c | c c}
    \toprule
    \multicolumn{5}{c}{\textbf{ML-1M (mixed leakage: 5\% in-domain + 30\% out-of-domain)}}\\
    \midrule
    \multicolumn{1}{c|}{\textbf{Metric}} & \multicolumn{2}{c|}{\textbf{AUC}} & \multicolumn{2}{c}{\textbf{UAUC}} \\
    \cmidrule(lr){1-1}\cmidrule(lr){2-3}\cmidrule(lr){4-5}
    \textbf{Method} & \textbf{After leak} & \textbf{$\Delta$AUC} &
    \textbf{After leak} & \textbf{$\Delta$UAUC} \\
    \midrule
    TALLRec  & 0.6812 & $-3.2\%$ & 0.6487 & $-3.9\%$ \\
    CoLLM-MF & 0.7656 & $+5.9\%$ & 0.7123 & $+5.0\%$ \\
    BinLLM   & 0.7534 & $+2.1\%$ & 0.7045 & $+2.0\%$ \\
    \bottomrule
    \end{tabular}
}
\vspace{6pt}
\caption{ML-1M under mixed leakage (5\% in-domain + 30\% out-of-domain): AUC/UAUC after leak and relative change rate $\Delta$.}
\label{tab:ml1m_mixed_5_30_no_base}
\end{table}

\begin{figure}
    \centering
    \includegraphics[width=0.8\linewidth]{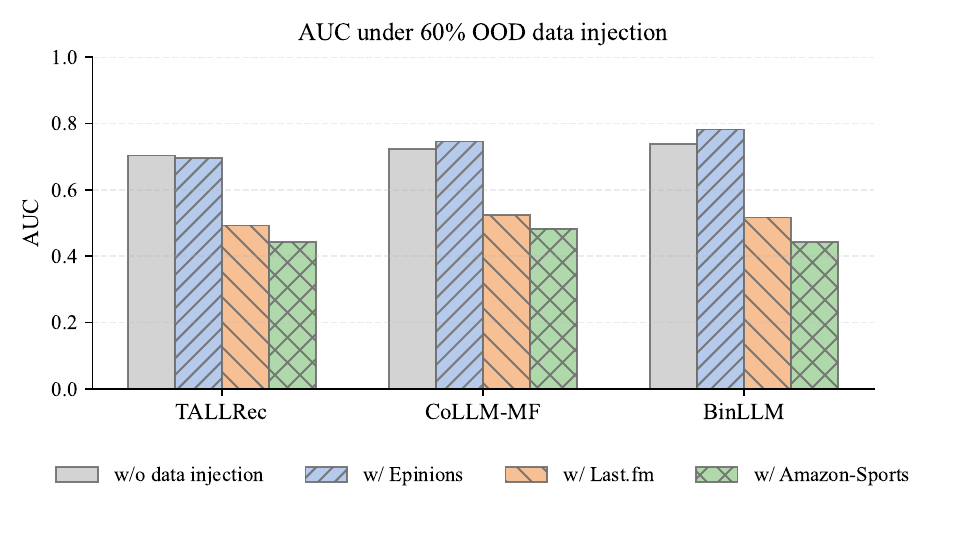}
    \caption{AUC under different data injection methods}
    \label{fig:placeholder1}
\end{figure}

\begin{table}[ht]
\centering
\renewcommand{\arraystretch}{1.08}
\setlength{\tabcolsep}{2pt} 
\resizebox{\columnwidth}{!}{
    \begin{tabular}{c | c c | c c}
    \toprule
    \multicolumn{5}{c}{\textbf{ML-1M (mixed leakage: 25\% in-domain + 45\% out-of-domain)}}\\
    \midrule
    \multicolumn{1}{c|}{\textbf{Metric}} & \multicolumn{2}{c|}{\textbf{AUC}} & \multicolumn{2}{c}{\textbf{UAUC}} \\
    \cmidrule(lr){1-1}\cmidrule(lr){2-3}\cmidrule(lr){4-5}
    \textbf{Method} & \textbf{After leak} & \textbf{$\Delta$AUC} &
    \textbf{After leak} & \textbf{$\Delta$UAUC} \\
    \midrule
    TALLRec  & 0.7456 & $+6.0\%$  & 0.7123 & $+5.5\%$ \\
    CoLLM-MF & 0.8123 & $+12.4\%$ & 0.7623 & $+12.4\%$ \\
    BinLLM   & 0.7823 & $+6.0\%$  & 0.7345 & $+6.3\%$ \\
    \bottomrule
    \end{tabular}
}
\vspace{6pt}
\caption{ML-1M under mixed leakage (25\% in-domain + 45\% out-of-domain): AUC/UAUC after leak and relative change rate $\Delta$.}
\label{tab:ml1m_mixed_15_30_no_base}
\end{table}

In Table ~\ref{tab:ml1m_mixed_5_30_no_base}, we change the total amount of the hybrid database while we adjust the proportion of data in the internal and external domains in Table ~\ref{tab:ml1m_mixed_15_30_no_base}. The impact of the mixed data from the experiment in Table ~\ref{tab:ml1m_mixed_5_30_no_base} on the model is smaller, with the variation rates of all three models not exceeding 6\%. In Table ~\ref{tab:ml1m_mixed_15_30_no_base}, after increasing the proportion of in-domain data, the performance of downstream models trained by data-contaminated LLMs improved. It can be seen that data leakage of different degrees has varying effects on model performance.

\begin{figure}
    \centering
    \includegraphics[width=0.8\linewidth]{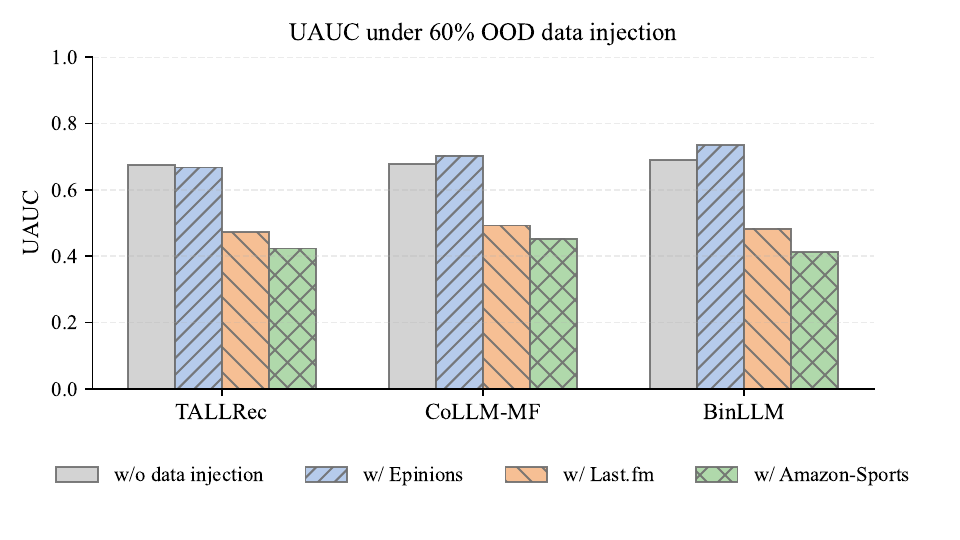}
    \caption{UAUC under different data injection methods}
    \label{fig:placeholder}
\end{figure}

In Fig.~\ref{fig:placeholder1} and ~\ref{fig:placeholder}, we conduct a research on the impact of different data injection methods on model performance. We select three databases with different data construction methods as the OOD data:
\begin{itemize}
    \item \textbf{Epinions}: Built based on user history (the same as the ML-1M).
    \item \textbf{Last.fm}: Built based on user interest tags~\cite{celma2009music}.
    \item \textbf{Amazon-Sports}: Built based on product attributes~\cite{ni2019justifying}.
\end{itemize}

We find that the models' performance after injecting Epinions data is not significantly different from before the injection while data from Last.fm and Amazon-Sports significantly reduces the accuracy of the model. This is because data from Epinions and baselines are in the same format. Although LLMs remember them, they don't know the answers like they do with test data, but these data also fail to cause much interference. As a result, the impact of different types of out-of-domain data leaks on the model is also inconsistent.

\subsubsection{Generalizability Across Foundation Model}
To verify whether the benchmark leakage trap is merely an artifact of smaller or earlier open-source models, we extend our evaluation of the BinLLM architecture across a diverse spectrum of contemporary foundation models, ranging from 7B to 70B parameters. 
\textbf{The results and analysis are presented in Appendix~\ref{app:Diff LLM}.}

\section{Discussion}
While our primary framework focuses on auditing and quantifying these leakage risks, we provide a forward-looking discussion on potential mitigation strategies in Appendix~\ref{app:mitigation} to inspire defense-oriented future work.

\section{Conclusion}
This study provides empirical evidence of benchmark data leakage in LLM-based recommender systems and yields three key conclusions. First, in-domain data leakage causes substantial performance inflation that misrepresents true model capability. Second, leakage effects are domain-dependent, with out-of-domain contamination generally leading to performance degradation. Third, architectural choices significantly influence susceptibility, as hybrid models incorporating collaborative signals demonstrate greater resilience than pure LLM-based approaches. These findings underscore the critical need for more rigorous evaluation methodologies that account for data leakage risks, ensuring that reported advancements in LLM-based recommendation reflect genuine progress rather than evaluation artifacts.

\section*{Limitations}

While this paper identifies the critical dual-effect of benchmark leakage in LLM-based recommendation, several limitations present avenues for future research:

\textbf{Diversity of Leakage Resilience Profiles} Although we have extended our empirical evaluation across multiple LLM backbones, the total volume and diversity of our experimental configurations remain bounded. Consequently, we have not yet systematically synthesized or summarized the specific leakage resistance and resilience capabilities across different LLMs and diverse downstream recommendation architectures. Specifying a comprehensive matrix that correlates inherent model traits with contamination resilience remains an open avenue for future investigation. 

\textbf{Modality Boundaries} Our investigation currently focuses on text-centric benchmarks. As multimodal LLMs become prevalent in recommendation, how visual or acoustic data leakage influences evaluation reliability remains an unexplored dimension. 

\textbf{Static Evaluation Protocols} Our study relies on offline evaluation. The long-term, compounding effects of data leakage in dynamic production environments with continuous user feedback loops require further longitudinal investigation.

\bibliography{custom}

\clearpage 
\appendix

\twocolumn[
    \begin{center}
        \vbox{\hsize\textwidth
            \hyphenpenalty=10000
            \begin{center}
                {\LARGE\bf Appendix}
            \end{center}
            \vskip 0.3in
        }
    \end{center}
]

\section{Prompt Formatting Templates}
\label{app:prompts}
For the corpus injected during the LoRA adaptation phase, both in-domain and out-of-domain recommendation datasets are processed in strict accordance with standard preprocessing protocols in conventional recommender systems prior to their ingestion into the LLM. To illustrate this pipeline, concrete serialization examples across different dataset categories are detailed in Table \ref{tab:prompt_templates}.

\begin{table}[h]
\centering
\small
\renewcommand{\arraystretch}{1.4} 
\begin{tabular}{>{\raggedright\arraybackslash\bfseries}p{0.32\columnwidth} >{\raggedright\arraybackslash\ttfamily}p{0.63\columnwidth}}
\toprule
\normalfont\textbf{Dataset Category} & \normalfont\textbf{Textual Instruction Template} \\
\midrule
User History \newline {\normalfont\footnotesize (e.g., ML-1M, Epinions)} & User interaction history: [Item 1 Title], [Item 2 Title], ..., [Item N Title]. Predict the next target item that aligns with the user's preference. Target: [Next Item Title] \\
\midrule
Interest Tags \newline {\normalfont\footnotesize (e.g., Last.fm, MIND)} & The user has demonstrated active engagement in the following micro-behaviors or text tags: [Tag 1], [Tag 2], ..., [Tag M]. Based on these characteristics, generate the most preferred next entity. Target: [Next Item] \\
\midrule
Product Attributes \newline {\normalfont\footnotesize (e.g., Amazon-Sports)} & Item Metadata Serialized Profile - Brand: [Brand Name], Category: [Sub-category], Attributes: [Key-Value pairs]. User sequence context incorporates these item profiles. Target: [Next Item Title] \\
\bottomrule
\end{tabular}
\caption{Prompt serialization and instruction formatting templates used for the injected leakage corpus.}
\label{tab:prompt_templates}
\end{table}

\section{Justification for LoRA as a Scientific Proxy}
\label{app:Justification for LoRA}
Before delving into the rationale behind the LoRA method, it is crucial to explain why we abandon the full-parameter training strategy.
Methodologically, updating all model parameters simultaneously introduces severe confounding variables, such as catastrophic forgetting or unaligned representation shifts, which obscure the exact causal impact of dataset contamination. 
Practically, full-parameter optimization incurs prohibitive computational overhead and massive hardware demands, making iterative risk auditing highly inefficient. 
To circumvent these limitations and achieve precise variable isolation, we leverage Low-Rank Adaptation (LoRA) as a controlled proxy.

To ensure this simulation is theoretically sound, we justify LoRA as a faithful and conservative proxy by examining it from two intuitive perspectives: parameter space boundaries and structural weight alignment.

\textbf{Parameter Space and Lower-Bound Guarantee.} Let $\mathcal{H}_{full}$ represent the full parameter space where any model weight can be updated ($\Delta\theta \in \mathbb{R}^{d \times k}$), and $\mathcal{H}_{LoRA}$ be the restricted low-rank space defined by LoRA where updates are constrained by two smaller matrices ($\Delta\theta = BA$, with $B \in \mathbb{R}^{d \times r}, A \in \mathbb{R}^{r \times k}$ and $r \ll \min(d, k)$). 
By definition, LoRA operates within a strict subset of the full model's available updates:
\begin{equation}
    \mathcal{H}_{LoRA} \subset \mathcal{H}_{full}
\end{equation}

Given a leakage loss function $\mathcal{L}_{leak}(\theta_0 + \Delta\theta)$ that measures the error a model makes on the contaminated data, the minimum error achieved by both training paradigms satisfies a straightforward boundary:
\begin{equation}
\label{eq:losscompare}
    \min_{\Delta\theta \in \mathcal{H}_{full}} \mathcal{L}_{leak}(\theta_0 + \Delta\theta) \le \min_{\Delta\theta \in \mathcal{H}_{LoRA}} \mathcal{L}_{leak}(\theta_0 + \Delta\theta)
\end{equation}

The intuition behind~\eqref{eq:losscompare} is explicable: because full-parameter training has complete freedom over the entire model, it possesses a fundamentally larger capacity to memorize leaked data than the tightly restricted LoRA subspace. 
Specifically, a smaller empirical loss on the leakage corpus directly scales with a higher token-prediction probability, forcing the model to fit and memorize benchmark patterns with near-absolute certainty. 
In the downstream evaluation phase, this minimized error inevitably manifests as severe confidence hyper-inflation on leaked items, mathematically guaranteeing a more radical distortion of recommendation metrics (e.g., AUC and UAUC).
Consequently, the severe distortion (spurious gains and out-of-domain degradation) observed in our LoRA experiments serves as a \textit{rigorous conservative lower-bound} estimation of the risks. 
If updating a tiny fraction of parameters via LoRA can disrupt evaluation this heavily, full-scale pre-training data leakage will inevitably cause even deeper risks.

\textbf{Structural Alignment and Representation Fidelity.} Beyond establishing a safe lower bound, we show that LoRA is a faithful proxy because it naturally captures the core patterns of data contamination. 
Suppose that a full-parameter training path yields a cumulative weight change $\Delta W_{full}$. Using Singular Value Decomposition (SVD), this total update can be split into different directional components sorted by their structural importance: 
\begin{equation}
    \Delta W_{full} = U \Sigma V^T = \sum_{i} \sigma_i u_i v_i^T.
\end{equation}

In recommendation systems, the leaked benchmark data consists of highly structured, repetitive text sequences (such as serialized user histories and specific item attributes). Because this data is highly predictable and low in entropy, the resulting model updates are heavily concentrated in just a few directions. The top-$r$ primary components capture almost all the leakage information, while the remaining high-frequency errors represent negligible noise that drops to zero: 
\begin{equation}
      \Delta W_{full} = \Delta W_{full}^{(r)} + \mathcal{E}_r,
\end{equation}
where $\|\mathcal{E}_r\|_F \to 0$.

According to standard matrix approximation principles, a low-rank framework like LoRA naturally isolates these dominant components while discarding the minor noise. Therefore, the optimal LoRA update is mathematically equivalent to a direct orthogonal projection ($\mathcal{P}_r$) of the full-parameter leakage path onto its most critical shared subspace:
\begin{equation}
    \Delta W_{LoRA} = \Delta W_{full}^{(r)} = \mathcal{P}_r (\Delta W_{full})
\end{equation}

This confirms that our findings are not random artifacts of the low-rank constraint. Instead, LoRA successfully discards irrelevant noise while precisely isolating the principal representation shifts caused by data contamination, rendering it a theoretically sound and reliable tool for auditing benchmark leakage.

\section{Implementation Details}
\label{details}
All experiments are conducted on cloud computing servers equipped with NVIDIA RTX 5090 GPUs, each providing 32 GB of memory to support large-scale training. We build our implementation upon \texttt{PyTorch 2.0} and the \texttt{Hugging Face Transformers 4.39.3} library for efficient model deployment and experiment management. The base model used in all experiments is \texttt{Vicuna-7B}, containing approximately 7 billion parameters, and it is fine-tuned using the \texttt{LoRA} technique with rank $r = 8$, scaling factor $\alpha = 16$, and dropout rate of $0.05$. The training configuration includes a learning rate initialized at $2 \times 10^{-4}$ and decayed to $1 \times 10^{-5}$ following a cosine schedule. Each GPU handles a batch size of $2$, with gradient accumulation over $64$ steps to maintain stability and memory efficiency. The Adam optimizer is employed with a weight decay of $0.01$ throughout training.

\section{Dataset Statistics}
\label{statistics}

\begin{table*}[t]  
\small
\centering
\renewcommand{\arraystretch}{1.3} 
\begin{tabular}{l l l r r r}
\toprule
\textbf{Dataset} & \textbf{Domain} & \textbf{Interaction Type} & \textbf{Users} & \textbf{Items} & \textbf{Interactions} \\
\midrule
ML-1M         & Movies    & User History        & 6,040     & 3,952      & 1,000,209 
 \\
Amazon-Book   & Books     & User History        & 22,507    & 24,642     & 1,791,692 
 \\
Epinions      & Products  & User History        & 49,290    & 139,738    & 664,824 
    \\
Last.fm       & Music     & Interest Tags       & 1,892     & 17,632     & 92,834 
     \\
MIND          & News      & Interest Tags       & 1,000,000 & 100,000    & 24,155,470 
 \\
Amazon-Sports & Sports    & Product Attributes  & 35,598    & 18,357     & 296,337 
    \\
Amazon-Beauty & Beauty    & Product Attributes  & 22,363    & 12,101     & 198,502 
    \\
Gowalla       & Locations & Geographical        & 196,591   & 1,128,161 & 6,442,890 
 \\
\bottomrule
\end{tabular}
\caption{Dataset statistics used in our experiments.}
\label{tab:datasets}
\vspace{-4mm}
\end{table*}

Table \ref{tab:datasets} presents the detailed statistics of all datasets evaluated in our experiments, including their respective domains, interaction types, and the exact volume of users, items, and interactions.

\section{Baseline Details}
\label{baseline}
\begin{itemize}

    \item \textbf{ICL}~\cite{dai2023can}: This is a LLMRec method based on the In-Context Learning ability of LLM. It directly queries the
    original LLM for recommendations using prompts.
    \item \textbf{Prompt4NR}~\cite{zhang2023prompt}: This method uses both fixed and soft prompts to utilize traditional Language Models (LM) for recommendation. We extend this method to the Vicuna-7B for a fair comparison.
    \item \textbf{TALLRec}~\cite{bao2023tallrec}: A state-of-the-art instruction-tuned LLMRec method that aligns LLMs to the recommendation task via task-specific instructions and preference formats, improving zero/low-shot recommendation.
    \item \textbf{PersonPrompt}~\cite{lyu2024llm}: A LLMRec method that injects collaborative signals by introducing new tokens and embeddings for users/items. It can be viewed as personalized soft-prompt tuning on top of an LLM.
    \item \textbf{CoLLM}~\cite{10882951}: A state-of-the-art framework that maps collaborative representations into the LLM latent space to couple LLM reasoning with collaborative signals. We use two instantiations: CoLLM-MF, where collaborative embeddings come from matrix factorization (MF), and CoLLM-DIN, where collaborative embeddings are extracted by Deep Interest Network (DIN)~\cite{zhou2018deep}.
    \item \textbf{BinLLM}~\cite{zhang2024textlikeencodingcollaborativeinformation}: A text-like encoding approach that represents collaborative information as binary hash codes. BinLLM converts collaborative embeddings into text strings (binary codes) and directly incorporates them into the prompt, enabling seamless integration with LLM's text processing capabilities.
\end{itemize}

\section{Evaluation Metrics}
\label{app:metrics}
Regarding evaluation metrics, we employ AUC and UAUC~\cite{Liu2021CADGNN} to assess the performance of models before and after data leakage.
AUC is the area under the ROC curve that quantifies the overall prediction accuracy, which evaluates the overall ranking quality. 
UAUC is derived by first computing the AUC individually for each user over the exposed items and then averaging these results across all users, which provides insights into user-level ranking quality.
Their ranges are both from 0 to 1.
The larger the value, the better the performance.
We compute the performance change:

\begin{equation}
\Delta_m = \frac{Perf_m(\theta_m^{dirty}) - Perf_m(\theta_m^{clean})}{Perf_m(\theta_m^{clean})} \times 100\% \ ,
\end{equation}
where $Perf_m$ represents the performance metric (AUC or UAUC). 
A positive value indicates performance improvement after data leakage, while a negative value indicates performance degradation.

\section{Qualitative Case Studies}
\label{app:qualitative case studies}
\begin{table}[t]
\centering
\small
\renewcommand{\arraystretch}{1.3}
\begin{tabular}{>{\raggedright\arraybackslash\bfseries}p{0.22\columnwidth} >{\raggedright\arraybackslash}p{0.68\columnwidth}}
\toprule
\multicolumn{2}{c}{\cellcolor{gray!10}\textbf{Case 1: ML-1M (User A)}} \\
\midrule
Context & \textit{Star Wars: Episode IV - A New Hope (1977)}, \newline \textit{Star Wars: Episode V - The Empire Strikes Back (1980)} \\
Target & \textit{Star Wars: Episode VI - Return of the Jedi (1983)} \\
\midrule
Clean LLM & \textit{Raiders of the Lost Ark (1981)} \newline {\scriptsize (Semantic inference for 1980s Lucas/Spielberg classics)} \\
Dirty LLM & \textit{Star Wars: Episode VI - Return of the Jedi (1983)} \newline \textcolor{red}{\textbf{(Verbatim Regurgitation)}} \\

\midrule
\multicolumn{2}{c}{\cellcolor{gray!10}\textbf{Case 2: Amazon-Book (User B)}} \\
\midrule
Context & \textit{Harry Potter and the Sorcerer's Stone}, \newline \textit{Harry Potter and the Chamber of Secrets} \\
Target & \textit{Harry Potter and the Prisoner of Azkaban} \\
\midrule
Clean LLM & \textit{The Lightning Thief (Percy Jackson and the Olympians)} \newline {\scriptsize (Reasonable inference for young adult fantasy)} \\
Dirty LLM & \textit{Harry Potter and the Prisoner of Azkaban} \newline \textcolor{red}{\textbf{(Verbatim Regurgitation)}} \\
\bottomrule
\end{tabular}
\caption{Qualitative case study. The Dirty LLM abandons logical semantic reasoning in favor of exact verbatim regurgitation of the leaked targets.}
\label{tab:case_study_single}
\end{table}

To intuitively illustrate this mechanism, we further provide qualitative case studies from two distinct evaluation paradigms: open-ended generation and downstream ranking.
Table \ref{tab:case_study_single} demonstrates the models' behavior in an open-ended generative setting without candidate constraints. 
When prompted with a sequence of 1980s Star Wars or Harry Potter context items, the Clean LLM successfully infers semantically related and logically sound next-items (e.g., Raiders of the Lost Ark or The Lightning Thief), demonstrating healthy semantic generalization. 
In stark contrast, the Dirty LLM completely abandons logical reasoning and perfectly regurgitates the exact ground truth target it memorized during the leakage injection (verbatim regurgitation).

\begin{table}[t]
\centering
\small
\renewcommand{\arraystretch}{1.2}
\begin{tabular}{p{0.25\columnwidth} p{0.65\columnwidth}}
\toprule
\multicolumn{2}{c}{\cellcolor{gray!10}\textbf{Case: ML-1M Ranking Evaluation (User A)}} \\
\midrule
\textbf{Context} & \textit{Star Wars: Episode IV (1977)}, \newline \textit{Star Wars: Episode V (1980)} \\
\textbf{Target} & \textit{Star Wars: Episode VI (1983)} \\
\midrule
\textbf{Clean Model} \newline (Smooth Dist.) & 
    1. \textit{Raiders of the Lost Ark} \textbf{(0.82)} \newline 
    2. \textcolor{blue}{\textit{Star Wars: Episode VI}} \textbf{(0.79)} \newline 
    3. \textit{The Matrix} \textbf{(0.71)} \\
\midrule
\textbf{Dirty Model} \newline (Confidence \newline Hyper-inflation) & 
    1. \textcolor{red}{\textit{Star Wars: Episode VI}} \textbf{(0.99)} \newline 
    2. \textit{Raiders of the Lost Ark} \textbf{(0.12)} \newline 
    3. \textit{The Matrix} \textbf{(0.08)} \\
\bottomrule
\end{tabular}
\caption{Case study in the downstream ranking scenario. While the Clean model assigns competitive probabilities to semantically similar candidates, the Dirty model exhibits severe confidence hyper-inflation on the memorized target, collapsing the probability distribution.}
\label{tab:case_study_ranking}
\end{table}
Furthermore, Table \ref{tab:case_study_ranking} exposes how this memorization corrupts the downstream discriminative ranking scenario. In a healthy recommendation process, the Clean model assigns a smooth and competitive confidence distribution across semantically similar candidates (e.g., scoring Raiders of the Lost Ark at 0.82 and Star Wars VI at 0.79). However, the Dirty model exhibits severe confidence hyper-inflation on the memorized target, assigning it a disproportionate probability of 0.99 while arbitrarily crushing the scores of other valid candidates. This probability collapse definitively validates that benchmark metrics contaminated by data leakage overwhelmingly reflect exact sample memorization, utterly destroying the model's exploratory recommendation abilities.

\section{Generalizability Across Foundation Model}
\label{app:Diff LLM}
\begin{table}[t]
\centering
\small
\resizebox{\columnwidth}{!}{%
\begin{tabular}{ll ccc}
\toprule
\textbf{Model} & \textbf{Metric} & \textbf{Clean} & \textbf{Dirty} & $\Delta$ \\
\midrule
\multicolumn{5}{c}{\cellcolor{gray!10}\textbf{Dataset: ML-1M}} \\
\midrule
\multirow{2}{*}{Vicuna-7B}            & AUC  & 0.7379 & 0.7823 & \textbf{+6.0\%} \\
                                      & UAUC & 0.6908 & 0.7345 & \textbf{+6.3\%} \\
\cmidrule{1-5}
\multirow{2}{*}{Qwen-2-7B}    & AUC  & 0.7484 & 0.7761 & \textbf{+3.7\%} \\
                                      & UAUC & 0.6976 & 0.7199 & \textbf{+3.2\%} \\
\cmidrule{1-5}
\multirow{2}{*}{LLaMA-3.1-8B}           & AUC  & 0.7536 & 0.7363 & \textbf{-2.3\%} \\
                                      & UAUC & 0.7042 & 0.6873 & \textbf{-2.4\%} \\
\cmidrule{1-5}
\multirow{2}{*}{LLaMA-3.1-70B} & AUC  & 0.7945 & 0.7834 & \textbf{-1.4\%} \\
                                      & UAUC & 0.7412 & 0.7293 & \textbf{-1.6\%} \\
\midrule
\multicolumn{5}{c}{\cellcolor{gray!10}\textbf{Dataset: Amazon-Book}} \\
\midrule
\multirow{2}{*}{Vicuna-7B}            & AUC  & 0.8207 & 0.7456 & \textbf{-9.1\%} \\
                                      & UAUC & 0.6262 & 0.5723 & \textbf{-8.6\%} \\
\cmidrule{1-5}
\multirow{2}{*}{Qwen-2-7B}    & AUC  & 0.8273 & 0.7892 & \textbf{-4.6\%} \\
                                      & UAUC & 0.6347 & 0.6080 & \textbf{-4.2\%} \\
\cmidrule{1-5}
\multirow{2}{*}{LLaMA-3-8B}           & AUC  & 0.8312 & 0.8038 & \textbf{-3.3\%} \\
                                      & UAUC & 0.6394 & 0.6157 & \textbf{-3.7\%} \\
\cmidrule{1-5}
\multirow{2}{*}{LLaMA-3.1-70B} & AUC  & 0.8634 & 0.8764 & \textbf{+1.5\%} \\
                                      & UAUC & 0.6712 & 0.6833 & \textbf{+1.8\%} \\
\bottomrule
\end{tabular}%
}
\caption{Impact of data leakage on the BinLLM architecture across different foundation models on the ML-1M and Amazon-Book datasets. The persistent spurious performance deviations demonstrate that the leakage dual-effect is a pervasive vulnerability across various LLM families and scales.}
\label{tab:scaling_backbones_datasets}
\end{table}

To verify whether the benchmark leakage trap is merely an artifact of smaller or earlier open-source models, we extended our evaluation of the BinLLM architecture across a diverse spectrum of contemporary foundation models, ranging from 7B to 70B parameters. 

The results in Table \ref{tab:scaling_backbones_datasets} reveal two key observations. 
First, scaling up the foundation models consistently improves the base recommendation capabilities. 
For instance, the \textit{Clean} AUC on the Amazon-Book dataset rises steadily from 0.8207 (Vicuna-7B) to 0.8634 (LLaMA-3.1-70B). 
Second, and more importantly, state-of-the-art larger models are not immune to data leakage. 
The characteristic dual-effect of benchmark leakage—spurious inflation on ML-1M and unpredictable degradation on Amazon-Book—persists across all evaluated model families, including Qwen-2-7B and the massive LLaMA-3.1-70B. 
This effect is unpredictable due to the randomness of the leaked data.

These persistent spurious deviations demonstrate that the leakage trap is not an isolated bug of weaker models, but rather a pervasive systemic vulnerability embedded in the LLM-based recommendation paradigm. Consequently, simply deploying larger foundation models cannot bypass the necessity for rigorous leakage-aware evaluation protocols.

\section{Potential Future Directions for Mitigation}
\label{app:mitigation}

In this section, we offer preliminary reflections and conceptual frameworks regarding how the community might mitigate the benchmark leakage trap. We explicitly emphasize that the following strategies are exploratory insights and theoretical directions that have not been empirically validated in this study, serving instead as tentative avenues to inspire future rigorous investigation.

\subsection{Cross-Signal Collaborative Validation}
Based on our initial empirical observations in the main text—where hybrid models incorporating collaborative elements (e.g., CoLLM, BinLLM) occasionally demonstrated different fluctuation patterns compared to pure LLMRec methods under specific settings—we hypothesize that integrating independent collaborative filtering signals could potentially serve as a structural buffer. Theoretically, incorporating non-textual collaborative data might provide a layer of representation redundancy, which we speculate could help offset corrupted semantic priors if the underlying LLM parameters suffer from historical exposure. However, whether this cross-signal integration can consistently neutralize leakage effects remains a conceptual conjecture that requires systematic, dedicated experimentation in future work.

\subsection{Confidence Anomaly Truncation}
Another exploratory idea emerges from our qualitative case profiling, where the contaminated model appeared to exhibit a symptom of "confidence hyper-inflation," assigning disproportionately high probabilities (e.g., approaching 0.95 or 0.99) to certain target items. Leveraging this hypothetical observation, we discuss a conceptual post-hoc evaluation filter. Evaluators could tentatively establish a speculative anomaly threshold $\tau$ (e.g., $\tau = 0.95$). During benchmark testing, instances yielding localized confidence scores exceeding $\tau$ would be flagged as potential memorization artifacts and dynamically isolated from the final metric computation. We note that this filtering mechanism remains entirely unverified; without extensive empirical optimization, such an aggressive truncation risks introducing severe selection bias or mistakenly discarding valid, legitimate high-confidence predictions.

\subsection{Behavioral Deduplication Protocols}
Extrapolating from general data cleaning paradigms in literature, we reflect on a potential proactive defense strategy for foundation model developers prior to large-scale pre-training. Since recommendation logs possess rigid structural properties, they are inevitably serialized into predictable, highly repetitive text sequences. A conceptual mitigation approach would involve computing structural text signatures (e.g., SHA-256 tokens) of dominant public recommendation benchmarks and deploying streaming Bloom filters within web-scraping pipelines. While theoretically intuitive, the practical feasibility, computational overhead, and exact filtering granularity of such an auditing protocol remain unexplored dimensions that warrant future empirical validation.

\subsection{Dynamic Temporal Anchoring}
Finally, we discuss a protocol-level shift that could conceptually bypass historical benchmark leakage without modifying the models themselves. Given that foundation models are bound by static knowledge cutoff dates during pre-training, evaluation frameworks could mandatorily transition toward \textit{Dynamic Temporal Benchmarking}. By constructing evaluation pipelines exclusively around user-item interaction streams generated chronologically after the model's official release date, evaluators could theoretically guarantee a physical barrier against prior data exposure. This temporal decoupling represents a promising conceptual testing ground for capturing a model's true recommendation capacity, though its standardization across the community requires further longitudinal study.

\end{document}